\documentclass{article}

\usepackage[preprint]{neurips}

\usepackage[utf8]{inputenc}
\usepackage[T1]{fontenc}
\usepackage{lmodern}
\usepackage{amsmath,amssymb,amsfonts}
\usepackage{bm}
\usepackage{graphicx}
\usepackage{booktabs}
\usepackage{xcolor}
\usepackage{algorithm}
\usepackage{algpseudocode}
\usepackage{microtype}
\usepackage{url}
\usepackage[colorlinks=true,allcolors=blue!55!black]{hyperref}

\newcommand{\vocab}{\mathcal{V}}
\newcommand{\loss}{\mathcal{L}}
\newcommand{\real}{\mathbb{R}}

\title{Typhoon: Gradient-Based Task-Specific\\ Masking for Masked Language Models}

\author{%
  Bruce Changlong Xu\thanks{Equal contribution.}
  \hspace{0.4em} Muhammed Shahir Abdurrahman\footnotemark[1]
  \hspace{0.4em} Hashem Elezabi\footnotemark[1] \\
  Department of Computer Science, Stanford University \\
  \texttt{\{brucechanglongxu, shahir, hashem\}@cs.stanford.edu}
}

\begin{document}
\maketitle

\begin{abstract}
The choice of \emph{which} tokens to mask is an under-examined design decision
in masked language modeling (MLM). Standard pretraining masks tokens uniformly
at random, whereas content-aware masking schemes can improve downstream
transfer. We study masking as a \emph{task-adaptive} component of the
fine-tuning pipeline and introduce \textbf{Typhoon}, a masking strategy that
uses the gradient of the MLM loss with respect to one-hot token inputs to
estimate online how much each token type contributes to the objective. Typhoon
maintains an exponential moving average of per-token-type saliency and
calibrates these scores into a masking distribution whose expected masking rate
matches a target budget under a token-independence approximation. We formalize
the method and evaluate it against random masking and whole-word masking on two
GLUE tasks, MRPC and CoLA, across three BERT-family backbones (TinyBERT,
DistilBERT, and BERT-base) and five random seeds per configuration ($90$
training runs in total). Once seed variance is accounted for, no masking
strategy is reliably better than the others on these tasks: on MRPC the gap
between Typhoon and the best baseline stays within $0.004$ $F_1$, across all
twelve Typhoon comparisons no paired test reaches significance, and every
$95\%$ confidence interval contains zero. Inspecting Typhoon's learned
distribution explains why: budget-preserving min--max calibration compresses
the gradient ranking into a narrow band whose Kullback--Leibler divergence from
uniform random masking is $7\!\times\!10^{-4}$ nats per token, and widening the
calibration spread by an order of magnitude in KL still leaves downstream
accuracy unchanged within seed noise. Typhoon's apparent advantage in
single-run experiments does not survive multi-seed evaluation, showing that
gradient-based task-adaptive masking is competitive with, but does not
outperform, uniform random masking at this scale.
\end{abstract}

\section{Introduction}
\label{sec:intro}

Transformer-based language models~\citep{vaswani2017attention} have driven rapid
progress across natural language processing, from machine translation and
question answering to sequence classification. The standard recipe has two
stages: self-supervised pretraining on large unlabeled corpora, followed by
continued task-adaptive pretraining and supervised fine-tuning on a labeled
downstream dataset~\citep{devlin2019bert,liu2019roberta,gururangan2020dont}. In
\emph{masked} language models (MLMs) such as BERT~\citep{devlin2019bert}, a
subset of input tokens is corrupted and reconstructed from bidirectional
context to produce representations for natural language understanding.

In a standard MLM, special \texttt{[MASK]} tokens replace a uniformly random
$15\%$ of the input, and the model is trained to recover the hidden tokens from
their bidirectional context. The masking rule is usually treated as a fixed
hyperparameter, yet it determines what the model is forced to learn. Prior work
shows that the masking policy affects downstream transfer: higher masking rates
can improve representation quality~\citep{wettig2023mask}, and content-aware
schemes that mask spans, whole words, named entities, or statistically
correlated tokens can outperform uniform random masking on knowledge- and
reasoning-intensive tasks~\citep{joshi2020spanbert,cui2021wwm,sun2019ernie,levine2021pmi,sadeq2022informask}.
This raises the question of whether an effective masking distribution can be
adapted online from the target task itself during continued MLM pretraining.

We investigate this hypothesis with \textbf{Typhoon}, a masking strategy that
uses input-loss gradients~\citep{li2016visualizing} computed during training. The
underlying intuition is that tokens whose presence most affects the MLM loss are
those whose reconstruction is most informative, and should therefore be masked
preferentially. Typhoon computes the gradient of the loss with respect to the
\emph{one-hot} input representation, accumulates a per-token-type saliency score
with an exponential moving average for stability across batches, and converts
these scores into a masking distribution that respects a target masking budget.
Unlike whole-word, span, or entity masking, Typhoon requires no external
linguistic resources or parsers; unlike PMI-based masking, it adapts to the
training objective rather than to static corpus statistics alone.

Our contributions are:
\begin{itemize}
  \item We frame masking as a task-adaptive component of the fine-tuning
  pipeline and give a precise formulation of gradient-based, task-specific
  masking (\S\ref{sec:method}).
  \item We introduce \textbf{Typhoon}, an online algorithm that maintains
  EMA-smoothed token saliencies and calibrates them into a masking distribution
  with a controllable expected masking rate (Algorithms~\ref{alg:update}
  and~\ref{alg:calibrate}).
  \item We evaluate Typhoon against random and whole-word masking on MRPC and
  CoLA across three BERT-family backbones and five seeds per configuration, and
  find that with seed variance accounted for no strategy is reliably better
  than random masking---a negative, reproducibility-oriented result, backed by
  paired significance and equivalence tests, that tempers the optimistic
  reading of single-run experiments (\S\ref{sec:experiments}).
  \item We give a mechanistic explanation for the null: the budget-preserving
  calibration step squashes Typhoon's gradient-based ranking into a masking
  distribution operationally indistinguishable from uniform random masking
  (KL $\approx\!7\!\times\!10^{-4}$ nats per token,
  \S\ref{sec:mechanism}), pinpointing where any improved task-adaptive masking
  strategy will have to break the budget--spread bottleneck.
\end{itemize}

\section{Related Work}
\label{sec:related}

Masking strategies for MLMs span a spectrum from content-agnostic to
content-aware. We organize the most relevant lines of work below.

\paragraph{Random masking.}
The original BERT model masks a uniformly random $15\%$ of
tokens~\citep{devlin2019bert}, of which $80\%$ are replaced with
\texttt{[MASK]}, $10\%$ with a random token, and $10\%$ left unchanged. This
``80--10--10'' rule was long treated as canonical, but later analysis suggests
that simpler corruption can match or exceed it and that substantially higher
masking rates are often beneficial for large
models~\citep{liu2019roberta,wettig2023mask}. Random masking is a strong,
resource-free baseline, but it is agnostic to which tokens are worth predicting.

\paragraph{Whole-word masking.}
Subword tokenizers split many words into multiple pieces (e.g., \texttt{Jo
\#\#han \#\#sen}). Whole-word masking masks all subwords of a chosen word
jointly, preventing the model from trivially completing a word from its other
fragments and yielding consistent gains on several tasks~\citep{cui2021wwm}.

\paragraph{Span, entity, and PMI masking.}
SpanBERT masks contiguous spans to encourage modeling of longer-range
structure~\citep{joshi2020spanbert}; knowledge-integration methods such as ERNIE
mask named entities and phrases~\citep{sun2019ernie}; and PMI-masking selects
spans with high pointwise mutual information so that the masked content is
genuinely predictive rather than locally guessable~\citep{levine2021pmi}. More
recently, InforMask uses a PMI-based criterion to choose the most informative
tokens and improves factual recall (LAMA) and question answering
(SQuAD)~\citep{sadeq2022informask}. These methods share the premise that
\emph{what} is masked should be informative; they differ in the signal they use
to define informativeness. Typhoon uses the task gradient as that signal.

\paragraph{Learned and adaptive masking.}
Beyond fixed heuristics, masking can in principle be \emph{learned}. Our gradient
view is loosely analogous to AutoAugment in computer
vision~\citep{cubuk2019autoaugment}, which searches data-augmentation policies,
and to time-varying schedules that change the masking rate over the course of
training~\citep{wettig2023mask}. Finally, while ``80--10--10'' label corruption
was long assumed helpful, recent evidence suggests the random-token substitution
can be detrimental~\citep{wettig2023mask}; accordingly, we do not use it.

\section{Method}
\label{sec:method}

\subsection{Setup and notation}
Let $\vocab$ be the vocabulary with $|\vocab| = V$. An input is a token sequence
$\bm{x} = (x_1, \dots, x_n)$ with each $x_i \in \vocab$. We write
$\bm{e}_i \in \{0,1\}^V$ for the one-hot encoding of $x_i$ and
$\bm{E} \in \real^{V \times d}$ for the embedding matrix, so that the embedded
input is $\bm{E}^\top \bm{e}_i \in \real^d$. A masking operation selects a subset
$M \subseteq \{1,\dots,n\}$ of positions, replaces $\{x_i : i \in M\}$ with
\texttt{[MASK]}, and the MLM is trained to reconstruct them under loss $\loss$.
We target an \emph{expected masking rate} $p \in (0,1)$: in expectation a
fraction $p$ of tokens are masked.

\subsection{Gradient-based token saliency}
We seek a per-token-type score that reflects how strongly each token influences
the loss. Standard MLM implementations pass integer token ids to an embedding
lookup table, which is non-differentiable with respect to the discrete input.
Following input-gradient saliency methods in NLP~\citep{li2016visualizing}, we
multiply the one-hot representation $\bm{e}_i$ by the embedding matrix so that
the loss is differentiable with respect to $\bm{e}_i$. For each position we take
the partial derivative of the loss with respect to the one-hot coordinate of the
\emph{true} token occupying that position,
\begin{equation}
  s_i \;=\; \frac{\partial \loss}{\partial e_{i,\,t_i}},
  \qquad
  t_i =
  \begin{cases}
    x_i & \text{if position $i$ is unmasked},\\
    y_i & \text{if position $i$ is masked,}
  \end{cases}
  \label{eq:saliency}
\end{equation}
where $y_i$ is the gold token id at a masked position. This is a \emph{signed}
scalar attributed to the token type $t_i$. Reading the gradient at the true
token id (rather than the norm over the whole vocabulary) keeps the sign
information that the weight update below relies on.

\subsection{Token-type weights via exponential moving average}
A single batch gives a noisy estimate of token saliency, so we accumulate a
weight $w_v$ for every token type $v \in \vocab$ across batches. The sign of the
contribution depends on whether the token was masked in the input that produced
the gradient: empirically we found that \emph{adding} the saliency for masked
positions and \emph{subtracting} it for unmasked positions yields the most
useful weights. Intuitively, this rewards token types that the model could not
reconstruct well when hidden, and discounts those that were easy in context.

Within a batch we form a fresh per-type estimate by averaging the signed
saliencies separately over the masked and unmasked occurrences of each type and
summing the two,
\begin{equation}
  \hat{w}_v \;=\;
  \underbrace{\frac{1}{n^{\text{u}}_v}\!\!\sum_{i \in U_v} \big(-s_i\big)}_{\text{unmasked occurrences}}
  \;+\;
  \underbrace{\frac{1}{n^{\text{m}}_v}\!\!\sum_{i \in M_v} \big(+s_i\big)}_{\text{masked occurrences}},
  \label{eq:batchweight}
\end{equation}
where $U_v$ and $M_v$ are the unmasked (non-special) and masked positions of
type $v$ in the batch, with counts $n^{\text{u}}_v$ and $n^{\text{m}}_v$. The
running weight is then updated with an exponential moving average that places
weight $\lambda \in (0,1)$ on the new estimate,
\begin{equation}
  w_v \;\leftarrow\;
  (1-\lambda)\, w_v \;+\; \lambda\, \hat{w}_v .
  \label{eq:ema}
\end{equation}
The EMA stabilizes the weights across the many small, noisy per-batch updates.
Algorithm~\ref{alg:update} states the update in full.

\subsection{From weights to a calibrated masking distribution}
Weights are not probabilities: a large $w_v$ does not directly say how often type
$v$ should be masked, and naively masking high-weight types would violate the
target budget $p$. We convert weights to per-type masking probabilities
$\pi_v \in [0,1]$ with two requirements: (i) $\pi_v$ is monotonically increasing
in $w_v$, and (ii) the \emph{dataset-level} expected masking rate equals $p$.

To enforce (ii) we maintain a frequency estimate $f_v$, the relative frequency of
token type $v$ over the dataset seen so far, with $\sum_{v} f_v = 1$. We also
track per-sequence counts so that a token appearing multiple times in one
sequence is not double counted relative to its dataset frequency. Under a
token-independence approximation---each position is masked independently given
its type---the expected fraction of masked tokens is
$\sum_{v} f_v\, \pi_v$. Rather than an exponential map, Typhoon uses a simple
\emph{linear} (min--max) rescaling, which is what the original implementation
applies and which keeps the calibration interpretable. Let $\mathcal{S}$ be the
set of types with a nonzero accumulated weight. We first min--max normalize the
weights of $\mathcal{S}$ to $[0,1]$, then affinely map them into a band
$[1-\delta,\,1]$ of width $\delta = \texttt{dist\_spread\_frac}$, and finally
multiply by a single scalar $\beta$ chosen to hit the budget:
\begin{equation}
  \tilde{w}_v \;=\;
  \frac{w_v - \min_{u \in \mathcal{S}} w_u}{\max_{u \in \mathcal{S}} w_u - \min_{u \in \mathcal{S}} w_u},
  \qquad
  g_v \;=\; (1-\delta) + \delta\,\tilde{w}_v,
  \qquad
  \pi_v \;=\; \mathrm{clip}_{[0,1]}\!\big(\beta\, g_v\big),
  \label{eq:calibrate}
\end{equation}
with $\beta$ set so that $\sum_v f_v\, \pi_v = p$; when clipping is inactive this
gives $\beta = p / \sum_v f_v g_v$. The band $[1-\delta,1]$ bounds how far any
type's masking probability can be pushed away from the mean, so even the
lowest-weight token is still masked a nonzero fraction of the time. Token types
not yet observed receive the prior masking probability $p$ until enough
statistics accumulate. Algorithm~\ref{alg:calibrate} gives the procedure; full
pseudocode for both routines appears in Appendix~\ref{app:algorithms}.

\begin{algorithm}[t]
\caption{Typhoon weight update (per batch)}
\label{alg:update}
\begin{algorithmic}[1]
\Require batch $\bm{x}$, mask set $M$, gold ids $\bm{y}$, weights $\bm{w}\in\real^V$, decay $\lambda$
\State Encode $\bm{x}$ as one-hot $\{\bm{e}_i\}$, embed, and compute loss $\loss$
\State Backpropagate to obtain $\partial \loss / \partial \bm{e}_i$ for all $i$
\State $t_i \gets y_i$ if $i \in M$ else $x_i$ \Comment{true token id at each position}
\State $s_i \gets \partial \loss / \partial e_{i,t_i}$ \Comment{signed scalar saliency}
\For{each token type $v$ present in the batch}
  \State $\hat{w}_v \gets \mathrm{mean}_{i \in U_v}(-s_i) + \mathrm{mean}_{i \in M_v}(+s_i)$
  \State $w_v \gets (1-\lambda)\, w_v + \lambda\, \hat{w}_v$
\EndFor
\State \Return $\bm{w}$
\end{algorithmic}
\end{algorithm}

\begin{algorithm}[t]
\caption{Typhoon mask sampling (per sequence)}
\label{alg:calibrate}
\begin{algorithmic}[1]
\Require sequence $\bm{x}$, weights $\bm{w}$, frequencies $\bm{f}$, budget $p$, spread $\delta$
\State $\mathcal{S} \gets \{v : w_v \neq 0\}$
\State $\tilde{w}_v \gets (w_v - \min_{u\in\mathcal{S}} w_u)/(\max_{u\in\mathcal{S}} w_u - \min_{u\in\mathcal{S}} w_u)$ \Comment{min--max to $[0,1]$}
\State $g_v \gets (1-\delta) + \delta\,\tilde{w}_v$ \Comment{affine map to band $[1-\delta,1]$}
\State choose $\beta$ so that $\sum_v f_v\,\mathrm{clip}_{[0,1]}(\beta g_v) = p$
\State $\pi_v \gets \mathrm{clip}_{[0,1]}(\beta g_v)$ for all $v$
\State $M \gets \emptyset$
\For{each position $i$ in $\bm{x}$}
  \State with probability $\pi_{x_i}$, add $i$ to $M$ \Comment{unseen types use prior $p$}
\EndFor
\State \Return mask set $M$
\end{algorithmic}
\end{algorithm}

\subsection{Training pipeline}
Following the task-adaptive pretraining paradigm~\citep{gururangan2020dont}, we
study how an intermediate task-specific masked-training phase inserted before
supervised fine-tuning affects downstream performance. To verify that findings
are consistent across model scales, we run the same pipeline on three
BERT-family backbones: TinyBERT~\citep{jiao2020tinybert} (a $4$-layer distilled
model), DistilBERT~\citep{sanh2019distilbert}, and
BERT-base~\citep{devlin2019bert}. The pipeline is:
\begin{enumerate}
  \item Load pretrained backbone weights for masked language modeling.
  \item Continue MLM training on the target GLUE task's text, masking with the
  strategy under test (random, whole-word, or Typhoon). For Typhoon, interleave
  the weight update (Algorithm~\ref{alg:update}) with mask sampling
  (Algorithm~\ref{alg:calibrate}).
  \item Transfer the MLM encoder into a sequence-classification head.
  \item Fine-tune the classifier on the target task and evaluate.
\end{enumerate}

\section{Experiments}
\label{sec:experiments}

\subsection{Setup}
We evaluate on two GLUE tasks~\citep{wang2018glue} with complementary linguistic
demands: \textbf{MRPC}~\citep{dolan2005mrpc}, sentence-pair paraphrase detection,
and \textbf{CoLA}~\citep{warstadt2019cola}, single-sentence grammatical
acceptability. We compare three masking strategies during the intermediate MLM
phase: random masking, whole-word masking, and Typhoon. Each
(task, strategy, backbone) configuration is run with five random seeds, giving
$2 \times 3 \times 3 \times 5 = 90$ training runs in total; all runs use the
HuggingFace Transformers library~\citep{wolf2020transformers}, the AdamW
optimizer~\citep{loshchilov2019decoupled}, and identical learning-rate schedules
and masking budgets $p$. Following~\citet{wettig2023mask}, we omit the
``80--10--10'' random-token substitution. We report accuracy and $F_1$ for MRPC
and Matthews correlation alongside accuracy for CoLA, as mean $\pm$ standard
deviation over the five seeds. Full hyperparameters are listed in
Appendix~\ref{app:hparams}.

\subsection{Results}
\label{sec:results}
\begin{figure}[t]
\centering
\includegraphics[width=\linewidth]{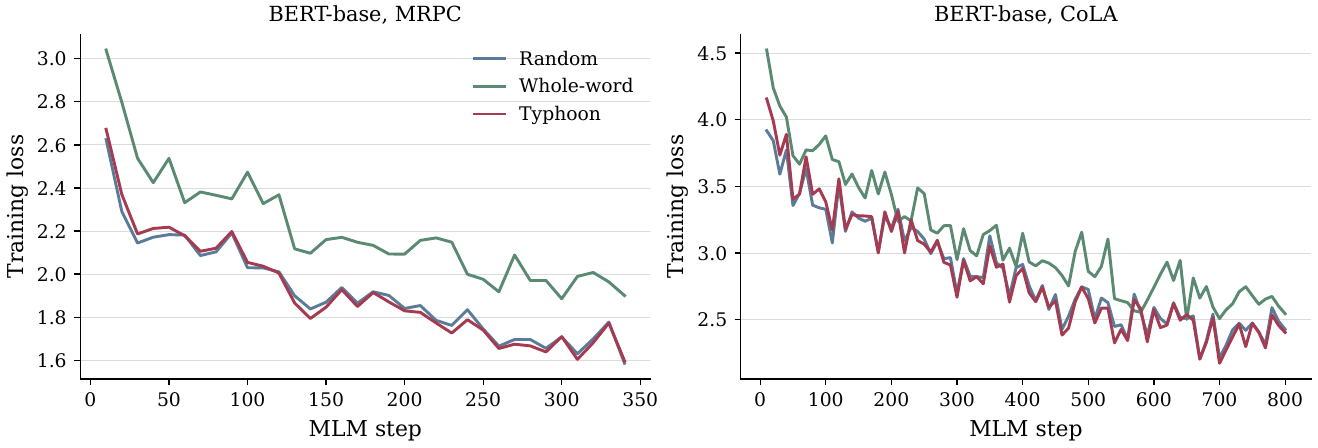}
\caption{Intermediate-phase MLM training loss versus optimizer step for
BERT-base (logged every ten steps, seed $0$). Random and Typhoon masking are
nearly indistinguishable; whole-word masking holds a higher loss throughout
because it hides every subword of a masked word, yet this does not change
downstream accuracy in Table~\ref{tab:main}.}
\label{fig:traj}
\end{figure}

\begin{figure}[t]
\centering
\includegraphics[width=\linewidth]{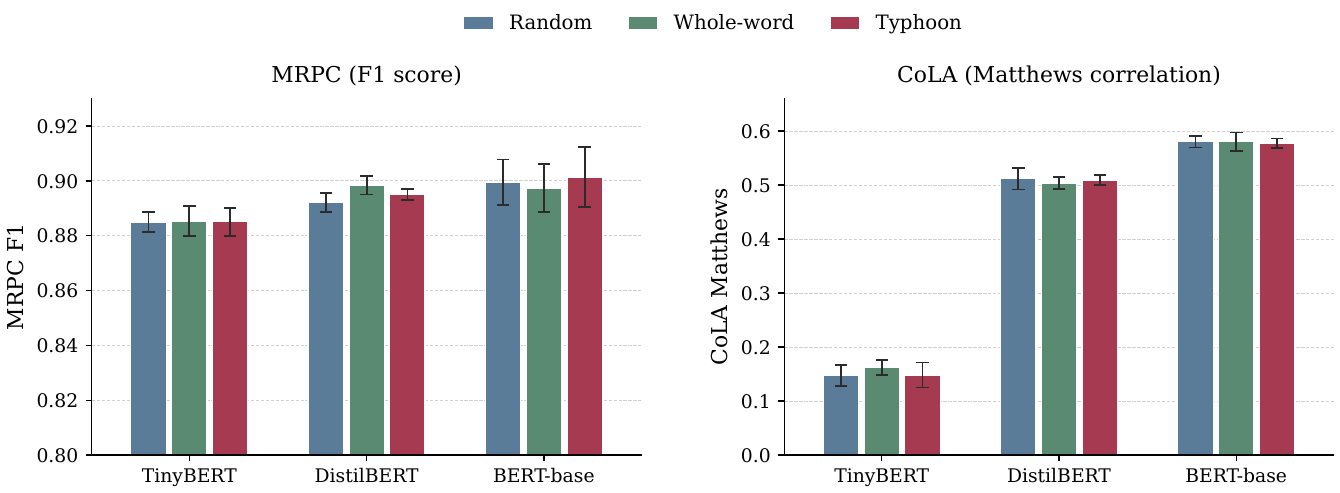}
\caption{Primary downstream metric by masking strategy and backbone (MRPC $F_1$,
CoLA Matthews correlation), mean over five seeds with error bars showing one
standard deviation. The strategies are within noise of each other at every
scale.}
\label{fig:bars}
\end{figure}

Table~\ref{tab:main} reports downstream performance for every backbone and
masking strategy, and Figure~\ref{fig:bars} visualizes the primary metric on
each task. During the intermediate MLM phase (Figure~\ref{fig:traj}) random and
Typhoon masking track nearly identical loss curves, while whole-word masking
sits at a consistently higher loss---masking all subwords of a word at once is a
harder reconstruction problem. This higher MLM loss does not translate into
worse (or better) downstream accuracy, which confirms that MLM loss alone is not
predictive of fine-tuned quality. In the
classification phase, validation metrics rose over the first few epochs and then
plateaued or declined as the models began to overfit; we report the best
validation checkpoint.

\begin{table}[t]
\centering
\caption{Downstream results on the GLUE validation sets across three backbones,
reported as mean $\pm$ standard deviation over five seeds. Within each
backbone--task block no strategy is reliably best on the primary metric; in
particular Typhoon stays within $0.004$ $F_1$ of the best baseline on MRPC.}
\label{tab:main}
\footnotesize
\setlength{\tabcolsep}{3pt}
\resizebox{\linewidth}{!}{% Auto-generated by scripts/collect_results.py. Do not edit by hand.
\begin{tabular}{llcccc}
\toprule
Backbone & Masking & MRPC Acc. & MRPC F1 & CoLA Matthews & CoLA Acc. \\
\midrule
TinyBERT & Random masking & 0.833\,$\pm$\,0.006 & 0.885\,$\pm$\,0.004 & 0.148\,$\pm$\,0.020 & 0.697\,$\pm$\,0.006 \\
 & Whole-word masking & 0.832\,$\pm$\,0.010 & 0.885\,$\pm$\,0.006 & 0.162\,$\pm$\,0.014 & 0.700\,$\pm$\,0.002 \\
 & Typhoon (ours) & 0.833\,$\pm$\,0.008 & 0.885\,$\pm$\,0.005 & 0.148\,$\pm$\,0.023 & 0.695\,$\pm$\,0.008 \\
\midrule
DistilBERT-base & Random masking & 0.845\,$\pm$\,0.004 & 0.892\,$\pm$\,0.003 & 0.512\,$\pm$\,0.020 & 0.799\,$\pm$\,0.010 \\
 & Whole-word masking & 0.852\,$\pm$\,0.003 & 0.899\,$\pm$\,0.003 & 0.504\,$\pm$\,0.011 & 0.798\,$\pm$\,0.004 \\
 & Typhoon (ours) & 0.849\,$\pm$\,0.003 & 0.895\,$\pm$\,0.002 & 0.509\,$\pm$\,0.009 & 0.798\,$\pm$\,0.006 \\
\midrule
BERT-base & Random masking & 0.855\,$\pm$\,0.011 & 0.900\,$\pm$\,0.008 & 0.580\,$\pm$\,0.011 & 0.828\,$\pm$\,0.004 \\
 & Whole-word masking & 0.853\,$\pm$\,0.012 & 0.897\,$\pm$\,0.009 & 0.581\,$\pm$\,0.017 & 0.829\,$\pm$\,0.006 \\
 & Typhoon (ours) & 0.857\,$\pm$\,0.017 & 0.901\,$\pm$\,0.011 & 0.577\,$\pm$\,0.009 & 0.827\,$\pm$\,0.003 \\
\bottomrule
\end{tabular}
}
\end{table}

Across all three backbones the three masking strategies land close to one
another on the primary metric. On \textbf{MRPC}, Typhoon edges out the baselines
for BERT-base ($0.901$ vs.\ $0.900$ $F_1$) but trails whole-word masking for
DistilBERT ($0.895$ vs.\ $0.899$); the largest gap between any two strategies
within a backbone is $0.007$ $F_1$ (DistilBERT), comparable to the per-strategy
seed standard deviation (up to $0.011$). On \textbf{CoLA}, whole-word masking is
nominally best for TinyBERT ($0.162$ vs.\ $0.148$ Matthews) and BERT-base
($0.581$ vs.\ $0.580$), while random masking leads for DistilBERT ($0.512$);
all of these gaps fall within overlapping error bars. We do not read any of
these differences as evidence that one strategy dominates.

\subsection{Statistical analysis}
\label{sec:stats}
Visual overlap of error bars is suggestive but not conclusive, so we test the
differences directly. Because every seed trains all three strategies for a given
backbone and task, the per-seed differences are matched, and we use a paired
$t$-test on the primary metric (full results in
Appendix~\ref{app:stats}, Table~\ref{tab:stats}). Across the $18$ pairwise
comparisons (three backbones, two tasks, three strategy pairs), exactly one
reaches significance at $\alpha = 0.05$: whole-word masking beats random masking
on DistilBERT MRPC ($\Delta = 0.006$ $F_1$, $p = 0.03$). None of the twelve
comparisons involving \textbf{Typhoon} is significant (all $p \ge 0.10$), and
every Typhoon-versus-baseline $95\%$ confidence interval contains zero; the
largest absolute mean difference anywhere is $0.015$.

A non-significant test does not by itself establish equivalence, so we also run
two one-sided tests (TOST) against a margin of one point on the primary metric
($\pm 0.01$). On MRPC, all six Typhoon-versus-baseline comparisons pass the
equivalence test ($p_{\text{eq}} < 0.05$): the strategies are statistically
equivalent within a one-point margin. On CoLA the equivalence test is
inconclusive---Matthews correlation has higher seed variance, and with five
seeds the TOST is underpowered, so we can neither separate the strategies nor
certify their equivalence at this margin. The combined picture is a clear null
for Typhoon on MRPC and an underpowered null on CoLA; in neither case does
Typhoon show an advantage.

A post hoc power analysis makes the task asymmetry explicit. Under a paired
$t$-test, detecting a true one-point effect at $80\%$ power requires a median of
only four seeds on MRPC (range $3$--$7$ across the comparisons) but about thirty
on CoLA (range $5$--$108$), because Matthews correlation is far noisier across
seeds at this dataset size. Our five seeds are therefore well powered to detect a
one-point effect on MRPC and underpowered on CoLA---the same split the
equivalence tests report---so the MRPC null is the load-bearing one.

\subsection{Compute cost}
\label{sec:cost}
Typhoon's saliency update needs a gradient with respect to the one-hot input on
top of the usual parameter gradient, so each intermediate-phase step costs more.
Table~\ref{tab:timing} gives the median wall-clock time per optimizer step on an
NVIDIA H100, measured over the full intermediate phase. Whole-word masking is
essentially free relative to random masking, whereas Typhoon adds $28$--$38\%$
per-step overhead, shrinking as the backbone grows because the extra
input-gradient work is a smaller fraction of a larger forward and backward pass.
The overhead falls only on the short intermediate phase, not on fine-tuning or
inference, but it is a real cost---and one a competitive method would need to
repay with a downstream gain that we do not observe.

\begin{table}[t]
\centering
\caption{Median wall-clock time per optimizer step during the intermediate MLM
phase (NVIDIA H100, batch size $32$, fp16), averaged over MRPC and CoLA. The
final column is Typhoon's per-step overhead relative to random masking.}
\label{tab:timing}
\small
% Auto-generated by scripts/make_benchmark_outputs.py. Do not edit by hand.
\begin{tabular}{lcccc}
\toprule
Backbone & Random & Whole-word & Typhoon & Overhead \\
 & (ms/step) & (ms/step) & (ms/step) & ($\times$ random) \\
\midrule
TinyBERT & 8.8 & 8.8 & 12.1 & 1.38 \\
DistilBERT & 13.5 & 13.5 & 18.4 & 1.37 \\
BERT-base & 20.5 & 20.5 & 26.3 & 1.28 \\
\bottomrule
\end{tabular}

\end{table}

\subsection{Why does Typhoon not help? A mechanistic look}
\label{sec:mechanism}
\begin{figure}[t]
\centering
\includegraphics[width=\linewidth]{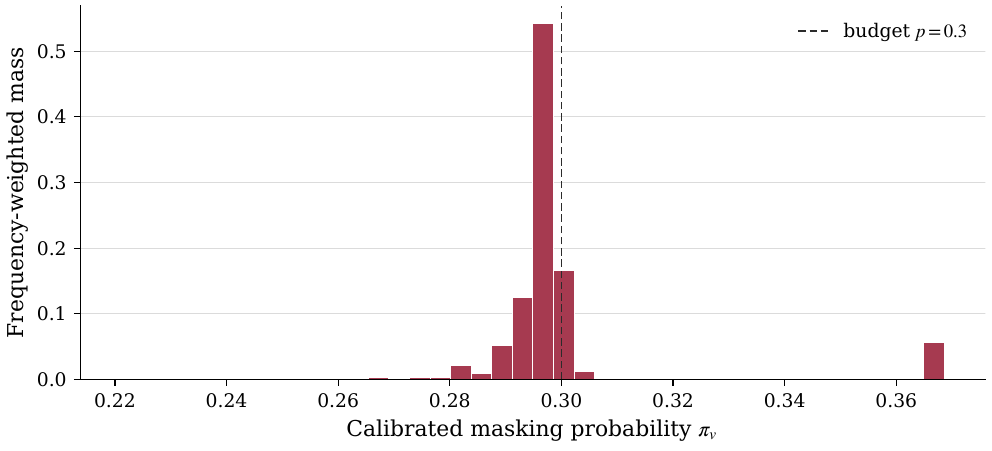}
\caption{Frequency-weighted distribution of Typhoon's calibrated per-type
masking probability $\pi_v$ after the intermediate MLM phase on BERT-base
(MRPC, seed $0$, default spread $\delta = 0.4$). Almost all token mass sits in
a narrow band of width $\approx 0.02$ around the budget $p = 0.3$, and the
per-token KL divergence against uniform random masking is $7\!\times\!10^{-4}$
nats---five orders of magnitude below $\log 2$. The budget-preserving
calibration squashes the gradient ranking into a distribution operationally
indistinguishable from random.}
\label{fig:rate_hist}
\end{figure}

The null result is more informative once we open up the algorithm. After three
epochs of intermediate MLM training on MRPC, BERT-base's Typhoon trainer has
seen $1.95\!\times\!10^5$ tokens and accumulated weights for $11{,}563$ of the
$30{,}522$ vocabulary entries. We compute the calibrated per-type masking
probabilities $\pi_v$ from these weights (Eq.~\ref{eq:calibrate}) and inspect
their distribution. Figure~\ref{fig:rate_hist} plots the frequency-weighted
histogram of $\pi_v$: under the default spread $\delta = 0.4$, the entire
distribution lives in the band $[0.22,\,0.37]$ around the budget $p = 0.3$,
with a frequency-weighted standard deviation of $0.017$.

Quantitatively, the calibrated masking distribution is almost indistinguishable
from uniform masking at rate $p$. Treating each token-type's masking decision as
a Bernoulli$(\pi_v)$, the per-token KL divergence against the uniform
Bernoulli$(p)$ is
\begin{equation*}
  \sum_v f_v\!\left[\pi_v \log\frac{\pi_v}{p} +
    (1-\pi_v)\log\frac{1-\pi_v}{1-p}\right]
  \;\approx\; 7\!\times\!10^{-4}\ \text{nats},
\end{equation*}
five orders of magnitude smaller than $\log 2$. In other words, the gradient
signal Typhoon collects is real but the budget-preserving min--max calibration
in Eq.~\eqref{eq:calibrate}---designed precisely to keep the expected masking
rate at $p$---compresses the per-type probabilities into a narrow band whose
distance from the random baseline is operationally negligible. This is the
mechanistic reason the downstream effect is also negligible.

The ranking itself is nevertheless informative: Typhoon's most-masked tokens are
predominantly rare proper-noun-like subwords (\texttt{neill}, \texttt{anatolia},
\texttt{dieter}, \texttt{vc}) for which the model's reconstruction loss is large
and noisy, while the least-masked types are common mid-frequency content words
(\texttt{transfer}, \texttt{fine}, \texttt{combined}, \texttt{out}) that the
model handles easily in context (Appendix~\ref{app:tokens},
Table~\ref{tab:tokens}). The signal is therefore consistent with the intuition
behind the method---tokens that are hard to reconstruct rise to the top---but
the calibration step is doing essentially all of the work in shaping the actual
masking policy.

We next test whether widening the spread parameter $\delta$ improves downstream
accuracy by allowing high-saliency tokens to deviate further from the budget
rate $p$. Table~\ref{tab:spread} reports this ablation: increasing $\delta$ from
the default $0.4$ to $0.8$ and $0.99$ widens the calibrated rate range from
$[0.22,0.37]$ to $[0.01,0.59]$ and increases the per-token KL divergence against
uniform masking by a factor of $17$, yet downstream MRPC accuracy and $F_1$ on
BERT-base remain within one seed standard deviation across all settings. Even at
$\delta = 0.99$, where Typhoon departs substantially from uniform random
masking, downstream performance is statistically indistinguishable from the
random baseline. At this corpus scale, per-token-type input-gradient saliency
does not identify a subset of tokens whose preferential masking improves
downstream fine-tuning.

\begin{table}[t]
\centering
\caption{Spread ablation on BERT-base / MRPC: varying Typhoon's spread $\delta$
moves the calibrated rate range and KL by more than an order of magnitude, but
downstream metrics stay flat within seed noise (three seeds per row).}
\label{tab:spread}
\small
% Auto-generated by scripts/make_spread_table.py. Do not edit by hand.
\begin{tabular}{lcccc}
\toprule
Spread $\delta$ & Rate range & KL (nats) & MRPC Acc. & MRPC $F_1$ \\
\midrule
$0.4$ (default) & [0.22, 0.37] & $7.0{\times}10^{-4}$ & 0.864\,$\pm$\,0.020 & 0.905\,$\pm$\,0.014 \\
$0.8$ & [0.10, 0.48] & $4.6{\times}10^{-3}$ & 0.851\,$\pm$\,0.015 & 0.897\,$\pm$\,0.010 \\
$0.99$ & [0.01, 0.59] & $1.2{\times}10^{-2}$ & 0.860\,$\pm$\,0.016 & 0.903\,$\pm$\,0.011 \\
\bottomrule
\end{tabular}

\end{table}

\subsection{Discussion}
Our conclusion is a null result: at this scale, gradient-based
task-adaptive masking neither helps nor hurts relative to random or whole-word
masking once seed variance is taken into account. This contrasts with the
single-run numbers from the original project, where Typhoon appeared to win on
MRPC---an apparent advantage that does not survive averaging over seeds and
backbones. The most likely explanation is that the intermediate MLM phase is
short and operates on small in-domain corpora (a few thousand sentences), so the
choice of \emph{which} tokens to mask has limited room to change the learned
representation before fine-tuning dominates. Two methodological points
generalize beyond this study: (i) masking-strategy comparisons on small GLUE
tasks are dominated by seed variance and require multiple seeds and backbones to
be meaningful, and (ii) reporting best-of-single-run results can manufacture
effects that vanish under repetition. We see value in publishing this corrected
picture rather than the original optimistic one.

\section{Limitations}
\label{sec:limitations}
Typhoon makes a token-independence approximation when calibrating masking
probabilities; correlations between adjacent tokens (which span- and PMI-based
methods exploit directly) are ignored. The method also introduces extra
hyperparameters---the EMA decay $\lambda$, the spread $\delta$, and budget
$p$---and computing input gradients adds overhead relative to random masking.
Our spread ablation (Table~\ref{tab:spread}) varies $\delta$ from $0.4$ to
$0.99$ but holds $\lambda$ and $p$ at their original values; a fuller sweep is
plausible but unlikely to overturn the null on MRPC.
Most importantly, our evaluation is confined to two small GLUE tasks and three
BERT-family backbones up to BERT-base; the intermediate MLM phase is short and
in-domain. The null result we report may not hold for longer task-adaptive
pretraining, larger models, knowledge-intensive tasks, or higher masking
budgets, and should be read as evidence about this regime rather than a verdict
on gradient-based masking in general. The sign convention in
Eq.~\eqref{eq:batchweight} was chosen empirically and deserves a more principled
justification.

\section{Conclusion and Future Work}
\label{sec:conclusion}
We presented Typhoon, a gradient-based, task-specific masking strategy for masked
language models that turns the loss gradient with respect to one-hot inputs into
an EMA-smoothed, budget-calibrated masking distribution. In a controlled
evaluation across three BERT-family backbones, two GLUE tasks, and five seeds per
configuration, Typhoon performs on par with random and whole-word masking: the
differences between strategies are smaller than the variation across seeds. The
stronger single-run result reported in the original project does not replicate
under this protocol. We therefore offer the method together with a
negative finding and a clean, reproducible implementation. Whether task-adaptive
masking pays off likely requires a different regime---longer adaptive
pretraining, larger models, knowledge-intensive tasks, learned masking policies
in the spirit of AutoAugment~\citep{cubuk2019autoaugment}, or schemes that model
token correlations rather than assuming independence. The baseline established
here should help future work revisiting this question.

\bibliographystyle{plainnat}
\bibliography{references}

\appendix

\section{Full Algorithms}
\label{app:algorithms}
The two routines in \S\ref{sec:method} are run jointly during the intermediate
MLM phase. Algorithm~\ref{alg:update} updates token-type weights from the
gradients of each batch; Algorithm~\ref{alg:calibrate} samples the mask for each
sequence from the current weights subject to the budget. Frequencies $f_v$ are
updated online as sequences are observed, and per-sequence counts prevent a
token from being over-weighted because it happens to recur within one sequence.
Token types never yet observed are assigned the prior masking probability $p$.

\section{Statistical tests}
\label{app:stats}
Table~\ref{tab:stats} reports the paired comparisons summarized in
\S\ref{sec:stats}. For each backbone and task we test the two
Typhoon-versus-baseline differences on the primary metric (MRPC $F_1$, CoLA
Matthews correlation). $\Delta$ is the mean paired difference over the five
seeds, with its $95\%$ confidence interval; $p$ is the paired $t$-test
$p$-value (${}^{\ast}$ marks $p < 0.05$); and $p_{\text{eq}}$ is the TOST
equivalence $p$-value against a $\pm 0.01$ margin, with \checkmark{} indicating
statistical equivalence at $\alpha = 0.05$. No Typhoon comparison is
significant, and every MRPC comparison is equivalent within one point.

\begin{table}[!htbp]
\centering
\caption{Paired significance and equivalence tests for Typhoon against each
baseline on the primary metric (five seeds per row).}
\label{tab:stats}
\small
% Auto-generated by scripts/stats_tests.py. Do not edit by hand.
\begin{tabular}{llrrrr}
\toprule
Backbone & Comparison & $\Delta$ & 95\% CI & $p$ & $p_{\text{eq}}$ \\
\midrule
\multicolumn{6}{l}{\emph{MRPC (F1)}} \\
TinyBERT & Typhoon $-$ Random & $+0.0001$ & $[-0.0037, +0.0038]$ & $0.949$ & $0.001$\,\checkmark \\
TinyBERT & Typhoon $-$ Whole-word & $-0.0003$ & $[-0.0100, +0.0094]$ & $0.940$ & $0.025$\,\checkmark \\
DistilBERT & Typhoon $-$ Random & $+0.0030$ & $[-0.0009, +0.0070]$ & $0.102$ & $0.004$\,\checkmark \\
DistilBERT & Typhoon $-$ Whole-word & $-0.0033$ & $[-0.0095, +0.0028]$ & $0.206$ & $0.020$\,\checkmark \\
BERT-base & Typhoon $-$ Random & $+0.0017$ & $[-0.0035, +0.0070]$ & $0.405$ & $0.006$\,\checkmark \\
BERT-base & Typhoon $-$ Whole-word & $+0.0039$ & $[-0.0021, +0.0099]$ & $0.147$ & $0.024$\,\checkmark \\
\midrule
\multicolumn{6}{l}{\emph{CoLA (Matthews)}} \\
TinyBERT & Typhoon $-$ Random & $+0.0007$ & $[-0.0063, +0.0078]$ & $0.785$ & $0.011$\,\checkmark \\
TinyBERT & Typhoon $-$ Whole-word & $-0.0140$ & $[-0.0595, +0.0315]$ & $0.441$ & $0.590$\,-- \\
DistilBERT & Typhoon $-$ Random & $-0.0031$ & $[-0.0177, +0.0116]$ & $0.592$ & $0.129$\,-- \\
DistilBERT & Typhoon $-$ Whole-word & $+0.0051$ & $[-0.0063, +0.0165]$ & $0.280$ & $0.150$\,-- \\
BERT-base & Typhoon $-$ Random & $-0.0030$ & $[-0.0178, +0.0117]$ & $0.597$ & $0.130$\,-- \\
BERT-base & Typhoon $-$ Whole-word & $-0.0032$ & $[-0.0318, +0.0253]$ & $0.770$ & $0.273$\,-- \\
\bottomrule
\end{tabular}

\end{table}

\section{Most- and least-masked tokens under Typhoon}
\label{app:tokens}
Table~\ref{tab:tokens} lists the fifteen non-special tokens with the highest
and lowest calibrated masking probabilities $\pi_v$ after the intermediate
MLM phase on BERT-base (MRPC, seed $0$). The pattern is consistent across
seeds: high-ranked tokens are rare proper-noun-like subwords whose
reconstruction loss is large, while low-ranked tokens are common content
words the model handles easily.

\begin{table}[!htbp]
\centering
\caption{Typhoon's top and bottom token-type masking probabilities on
BERT-base / MRPC after three epochs of intermediate MLM training.}
\label{tab:tokens}
\small
% Auto-generated by scripts/analyze_typhoon_weights.py. Do not edit by hand.
\begin{tabular}{rlcrlc}
\toprule
 & \multicolumn{2}{c}{Most-masked tokens} & & \multicolumn{2}{c}{Least-masked tokens} \\
\cmidrule(lr){2-3}\cmidrule(lr){5-6}
Rank & Token & $\pi$ & Rank & Token & $\pi$ \\
\midrule
1 & \texttt{neill} & 0.369 & 1 & \texttt{\#\#ecure} & 0.221 \\
2 & \texttt{vc} & 0.345 & 2 & \texttt{implications} & 0.229 \\
3 & \texttt{\#\#dorf} & 0.343 & 3 & \texttt{transfer} & 0.235 \\
4 & \texttt{\#\#lvis} & 0.341 & 4 & \texttt{\#\#pac} & 0.245 \\
5 & \texttt{anatolia} & 0.329 & 5 & \texttt{fine} & 0.246 \\
6 & \texttt{leaf} & 0.324 & 6 & \texttt{combined} & 0.246 \\
7 & \texttt{50} & 0.324 & 7 & \texttt{out} & 0.253 \\
8 & \texttt{utter} & 0.323 & 8 & \texttt{lion} & 0.255 \\
9 & \texttt{\#\#40} & 0.322 & 9 & \texttt{violation} & 0.256 \\
10 & \texttt{especially} & 0.322 & 10 & \texttt{\#\#ction} & 0.257 \\
11 & \texttt{dieter} & 0.322 & 11 & \texttt{\#\#ata} & 0.257 \\
12 & \texttt{\#\#sha} & 0.321 & 12 & \texttt{\#\#bed} & 0.258 \\
13 & \texttt{neighbouring} & 0.320 & 13 & \texttt{reflection} & 0.258 \\
14 & \texttt{rr} & 0.320 & 14 & \texttt{assembly} & 0.258 \\
15 & \texttt{binding} & 0.320 & 15 & \texttt{\#\#alis} & 0.259 \\
\bottomrule
\end{tabular}

\end{table}

\section{Hyperparameters}
\label{app:hparams}
Table~\ref{tab:hparams} lists the key hyperparameters, shared across all three
backbones and both tasks.

\begin{table}[!htbp]
\centering
\caption{Hyperparameters for the intermediate MLM phase and fine-tuning.}
\label{tab:hparams}
\small
\setlength{\tabcolsep}{4pt}
\begin{tabular}{ll}
\toprule
Hyperparameter & Value \\
\midrule
Backbones & TinyBERT (4L/312d), DistilBERT-base, BERT-base \\
Masking budget $p$ & $0.3$ \\
EMA decay $\lambda$ (weight on new estimate) & $0.8$ \\
Probability spread $\delta$ & $0.4$ \\
Optimizer & AdamW \\
Learning rate (MLM / fine-tune) & $1\times10^{-4}$ / $2\times10^{-5}$ \\
Weight decay & $1\times10^{-2}$ \\
Batch size & $32$ \\
Precision & fp16 \\
MLM epochs / fine-tune epochs & $3$ / $5$ \\
Seeds per configuration & $5$ \\
\bottomrule
\end{tabular}
\end{table}

\end{document}